\documentclass[journal,twoside]{IEEEtran}

\IEEEoverridecommandlockouts                              % This command is only needed if 
\usepackage{epstopdf}
\usepackage{cite}
\usepackage{amsmath,amssymb,amsfonts}
\usepackage{algorithmic}
\usepackage{graphicx}
\usepackage{textcomp}
\usepackage{subcaption}
\usepackage[ruled,vlined]{algorithm2e}
\usepackage{float}
\usepackage{wrapfig}
\usepackage{xcolor}
\usepackage[inkscapelatex=false]{svg}
\usepackage{breqn}
\usepackage{url}

\usepackage{mathtools}
\usepackage{svg}
\usepackage[inkscapelatex=false]{svg}

\usepackage{pgffor}

\newtheorem{dfn}{Definition}

\newtheorem{thm}{Theorem}

\newtheorem{rem}{Remark}

\makeatletter
\newcommand{\subalign}[1]{%
	\vcenter{%
		\Let@ \restore@math@cr \default@tag
		\baselineskip\fontdimen10 \scriptfont\tw@
		\advance\baselineskip\fontdimen12 \scriptfont\tw@
		\lineskip\thr@@\fontdimen8 \scriptfont\thr@@
		\lineskiplimit\lineskip
		\ialign{\hfil$\m@th\scriptstyle##$&$\m@th\scriptstyle{}##$\crcr
			#1\crcr
		}%
	}
}

\newenvironment{prf}{\textbf{Proof.}}{\hfill$\blacksquare$}

\newcommand{\Tpred}{T_{\text{pred}}}									% Prediction horizon
\newcommand{\Treplay}{{T_\replay}}										% Replay buffer size
\newcommand{\lrcrit}{\alpha_\crit}										% Critic learning rate
\newcommand{\goaldist}[1][{}]{d_{\G#1}}									% Distance to goal set
\makeatletter
\newcommand{\pushright}[1]{\ifmeasuring@#1\else\omit\hfill$\displaystyle#1$\fi\ignorespaces}
\newcommand{\pushleft}[1]{\ifmeasuring@#1\else\omit$\displaystyle#1$\hfill\fi\ignorespaces}
\makeatother

\newcommand{\mquad}[1]{%
    \foreach \n in {1,...,#1} {%
        \qquad%
    }%
}

\newcommand{\alglinelabel}{%
  \addtocounter{ALC@line}{-1}% Reduce line counter by 1
  \refstepcounter{ALC@line}% Increment line counter with reference capability
  \label% Regular \label
}

\newcommand{\diff}{\mathop{}\!\mathrm{d}}								% Differential
\newcommand{\eps}{{\varepsilon}}										% Epsilon
\DeclareMathOperator*{\argmin}{\text{arg\,min}}							% Argmin
\newcommand{\inv}{\ensuremath{^{-1}}}									% Inverse
\newcommand{\nrm}[1]{\left\lVert#1\right\rVert}							% Norm
\newcommand{\moddiv}[2]{{#1}\,\text{mod}\,{#2}}							% Modulo division
\newcommand{\cl}[1]{\text{cl}\left({#1}\right)}							% Set closure
\newcommand{\low}{{\text{low}}}											% Lower bound
\newcommand{\up}{{\text{up}}}											% Upper bound
\newcommand{\ra}{\rightarrow}											% Right arrow
\newcommand{\la}{\leftarrow}											% Left arrow
\newcommand{\ie}{\unskip, i.\,e.,\xspace}								% That is
\newcommand{\eg}{\unskip, e.\,g.,\xspace}								% For example
\newcommand{\sut}{\text{s.\,t.\,}}										% Such that or subject to
\newcommand{\wrt}{w.\,r.\,t. \xspace}									% With respect to
\newcommand{\N}{{\mathbb{N}}}											% Set of natural numbers
\newcommand{\Z}{{\mathbb{Z}}}											% Set of integer numbers
\newcommand{\R}{{\mathbb{R}}}											% Set of real numbers
\newcommand{\blue}[1]{\textcolor{blue}{#1}}
\definecolor{dgreen}{rgb}{0.0, 0.5, 0.0}

\newcommand{\state}{s}													% State (as vector)
\newcommand{\states}{\mathbb S}											% State space
\newcommand{\action}{a}													% Action (as vector)	
\newcommand{\actions}{\mathbb A}										% Action space
\newcommand{\traj}{z}													% State-action tuple (as vector tuple)
\newcommand{\policy}{\pi}												% Policy (as function or distribution)
\newcommand{\policies}{\Pi}												% Policy space
\newcommand{\transit}{p}												% State transition map
\newcommand{\cost}{c}													% Cost (as vector)
\newcommand{\Value}{V}													% Value
\newcommand{\W}{\ensuremath{\mathbb{W}}}								% Weight space
\newcommand{\act}{{\text{act}}}											% Actor abbreviation
\newcommand{\crit}{{\text{crit}}}										% Critic abbreviation
\newcommand{\G}{\ensuremath{\mathbb{G}}}								% Attractor (goal set)
\newcommand{\K}{\ensuremath{\mathcal{K}}\xspace}						% Class kappa
\newcommand{\KL}{\ensuremath{\mathcal{KL}}\xspace}						% Class kappa-ell
\newcommand{\Kinf}{\ensuremath{\mathcal{K}_{\infty}}\xspace}			% Class kappa-infinity
\newcommand{\KLinf}{\ensuremath{\mathcal{KL}_{\infty}}\xspace}			% Class kappa-ell-infinity
\newcommand{\loss}{\mathcal L}											% Loss
\newcommand{\replay}{\mathcal R}										% Experience replay
\newcommand{\spc}{{\,\,}}												% White space to be used in logical formulas

\begin{document}

\title{Critic as Lyapunov function (CALF): a model-free, stability-ensuring agent}
%\author{Anonymous}
\author{Pavel Osinenko$^{1}$, Grigory Yaremenko$^{1}$, Roman Zashchitin$^{2}$, \\ Anton Bolychev$^{1}$, Sinan Ibrahim$^{1}$, Dmitrii Dobriborsci$^{2}$
\thanks{$^{1}$Skolkovo Institute of Science and Technology}
\thanks{$^{2}$Deggendorf Institute of Technology, Technology Campus Cham}
\thanks{Corresponding author: P. Osinenko, email: \texttt{p.osinenko@gmail.com}.
First two authors contributed equally.
Accepted for 27th IEEE Conference on Decision and Control.
}
%\thanks{Pavel Osinenko is the corresponding author.}
}

\IEEEoverridecommandlockouts

% \overrideIEEEmargins is no longer supported
\IEEEpubid{\makebox[\columnwidth]{978-1-4799-4937-3/14/\$31.00~\copyright{}2024 IEEE \hfill} \hspace{\columnsep}\makebox[\columnwidth]{ }}

\maketitle

%%%%%%%%%%%%%%%%%%%%%%%%%%%%%%%%%%%%%%%%%%%%%%%%%%%%%%%%%%%%%%%%%%%%%%%%%%%%%%%%%%%%%%%%%%%%
%%%%%%%%%%%%%%%%%%%%%%%%%%%%%%%%%%%%%%%%%%%%%%%%%%%%%%%%%%%%%%%%%%%%%%%%%%%%%%%%%%%%%%%%%%%%

\begin{abstract}
This work presents and showcases a novel reinforcement learning agent called Critic As Lyapunov Function (CALF) which is model-free and ensures online environment, in other words, dynamical system stabilization.
Online means that in each learning episode, the said environment is stabilized.
This, as demonstrated in a case study with a mobile robot simulator, greatly improves the overall learning performance.
The base actor-critic scheme of CALF is analogous to SARSA.
The latter did not show any success in reaching the target in our studies.
However, a modified version thereof, called SARSA-m here, did succeed in some learning scenarios.
Still, CALF greatly outperformed the said approach.
CALF was also demonstrated to improve a nominal stabilizer provided to it.
In summary, the presented agent may be considered a viable approach to fusing classical control with reinforcement learning.
Its concurrent approaches are mostly either offline or model-based, like, for instance, those that fuse model-predictive control into the agent.
\end{abstract}

%%%%%%%%%%%%%%%%%%%%%%%%%%%%%%%%%%%%%%%%%%%%%%%%%%%%%%%%%%%%%%%%%%%%%%%%%%%%%%%%%%%%%%%%%%%%
%%%%%%%%%%%%%%%%%%%%%%%%%%%%%%%%%%%%%%%%%%%%%%%%%%%%%%%%%%%%%%%%%%%%%%%%%%%%%%%%%%%%%%%%%%%%
\section{Background and problem statement}
\label{sec_problem}

\textbf{Notation:}
We will use Python-like array notation \eg $[0:T] = \{0, \dots, T-1\}$ or $\state_{0:T} = \{\state_0, \dots, \state_{T-1}\}$.
Spaces of class kappa, kappa-infinity, kappa-ell and kappa-ell-infinity functions are denoted $\K, \Kinf, \KL, \KLinf$, respectively.
These are scalar monotonically increasing functions, zero at zero, and, additionally, tending to infinity in case of $\Kinf$. 
The subscript $\ge 0$ in number set notation will indicate that only non-negative numbers are meant.
The notation ``$\cl{\bullet}$'', when referring to a set, will mean the closure.
%We treat $\deltau$ as a single symbol denoting the action step size \ie the physical time elapsed between each two consecutive $\action_t$ and $\action_{t+1}$.
We denote modulo division of the first argument by the second as ``$\moddiv{\bullet}{\bullet}$''.

\subsection{Problem statement}
Consider the following optimal control problem:

\begin{equation}
\label{eqn_ocproblem}
\begin{aligned}
	& \Value^\policy \left(\state_0 \right) = \sum_{t=0}^{\infty} \cost \left(\state_t, \action_t \right) \ra \min, & \\
	& \phantom{\sut \quad} \state_0 \in \states, \policy \in \policies, t \in \Z_{\ge 0} & \\
	& \sut \quad \state_{t+1} = \transit( \state_t, \action_t ), \action_t \in \actions(\state_t) & \\
	& \phantom{\sut \quad} \action_{t} = \policy(\state_t), &
\end{aligned}
\end{equation}

where:

\begin{enumerate}
\item $\states$ is the \textit{state space} \eg $\R^n, n \in \N$, that is a normed vector space of all states of the given environment;
\item $\actions(\state)$ is the \textit{action space}, in general state-dependent \eg a subset of $\R^m, m \in \N$, that is a set of all actions available to the agent at the current state.
It is assumed that $\actions(\state)$ is compact for each state $\state$;
%\item $\transit : \states \times \actions \times \states \ \rightarrow \ \R$ is the \textit{transition probability density function} of the environment, that is such function that $\transit(\cdot \mid \state_{t}, \action_{t})$ is the probability density of the next state $s_{t + 1}$ conditioned on the current state $\state_{t}$ and current action $\action_{t}$;
\item $\transit : \states \times \actions \ra \states$ is the \textit{state transition map} of the environment, assumed upper semi-continuous in norm;
\item $\cost : \states \times \actions \rightarrow \mathbb{R}$ is the \textit{cost function} of the problem, that is a function that takes a state $\state_{t}$ and an action $\action_{t}$ and returns the immediate cost $\cost_{t}$ incurred upon the agent if it were to perform action $\action_{t}$ while in state $\state_{t}$;
\item $\policies$ is the \textit{set of policies} (state feedback control laws in other words), that is a set of functions $\policy : \states \ra \actions$.
\end{enumerate}

Sometimes, also a discount factor $\gamma \in (0, 1)$ is introduced in front of the cost in \eqref{eqn_ocproblem} as a factor $\gamma^t$.

The goal of reinforcement learning is to find a policy $\policy$ that minimizes $V^{\policy}(\state_0)$ in \eqref{eqn_ocproblem}.
The policy $\policy^*$ that solves this problem is commonly referred to as \textit{the optimal policy}.
An \textit{agent} will be referred to as a synonym for $\policy$ which was generated by some finite algorithm \eg actor-critic.
In the latter, for instance, one step of the algorithm tries to approximate \ie ``learn", the value $\Value^\policy$ via a neural network $\hat \Value^w$ with weights $w$ based on some collected data called \textit{replay buffer}, say, $\replay = \{\state_{t_k}, \action_{t_k}\}_{k \in [0:\Treplay]}$ of size $\Treplay$ and \sut $t_k < t_{k+1}, \forall k$.
In the second step, the algorithm may choose a so-called \textit{greedy} or \eg $\eps$-greedy action $a_t$, which minimize an actor loss $\loss^\act$ \eg $\loss^\act(\action_t) = \cost_t + \hat \Value^{w_t}( \state_{t+1} )$ or take a random action with probability $\eps>0$, respectively. 
A useful object is the action-value or Q-function $Q^\policy(\state, \action) \triangleq \cost(\state, \action) + \Value^\policy( \state_+^\action )$, where $\state_+^\action$ is the state at the next step relative to $\state$ upon taking the action $\action$.
The herein present approach, CALF, will base particularly on Q-function.

Common reinforcement learning approaches to the above include various methods of policy gradient, actor-critics and tabular dynamic programming (see \eg \cite{Sutton2018ReinforcementL,Kakade2001naturalpolicy,Baxter2001Infinitehorizo,Peters2006Policygradient,Bertsekas2019Reinforcementl}).
A common classical control approach is in turn \eg model-predictive control (MPC) that places some finite horizon $\Tpred > 2$ in place of the infinity in \eqref{eqn_ocproblem} and optimizes over respective sequences of actions via state propagation through the model $\transit$.
A \textit{model-free} agent, as will be understood in this work, is a controller that does not use $\transit$, or any learned model thereof, to compute actions.
Such an agent is also called \textit{data-driven} for instance.
Notice, not every reinforcement learning agent is model-free \eg Dyna \cite{Sutton1991Dynaintegrated,Pei2021improveddynaq}.

Let now $\G \subset \states, 0 \in \G$ denote a \textit{goal set} to which we want to drive the state optimally according to \eqref{eqn_ocproblem}.
%Let the cost $\cost$ be continuous, strictly positive outside $\G$ and diverging with the distance to $\G$.
We assume, $\G$ to be a compact neighborhood of the origin.
Also a compact in a subspace of the state space spanned by some state variables of interest may be considered, but we omit this case for simplicity.
Let $d_\G(\state) := \inf\limits_{\state' \in \G} \nrm{\state - \state'}$ be the distance to the goal.
We call a policy $\policy_0 \in \policies$ a \textit{$\G$-stabilizer} if setting  $a_{t} = \policy_0(\state_{t}), \forall t$ implies that the distance between $\state_{t}$ and $\G$ tends to zero over time.
Formally, we use the following definition.

\begin{dfn}
	\label{dfn_stabilizer}
	A policy $\policy_0 \in \policies$ is called a \textit{$\G$-stabilizer}, or simply a \textit{stabilizer}, if the goal set is implied by the context, if
	\begin{equation}
		\label{eqn_goallim}
		\forall t \ge 0 \spc \action_t \gets \policy_0(\state_t) \implies \forall \state_0 \in \states \spc \lim\limits_{t \ra \infty}\goaldist(\state_t) = 0.
	\end{equation}	
	It is called a uniform $\G$-stabilizer if, additionally, the limit in \eqref{eqn_goallim} is compact-convergent \wrt $\state_0$ and
	\begin{equation}
		\label{eqn_goalstab}
		\begin{aligned}
			& \exists \eps_0 \ge 0 \spc \forall t \ge 0 \spc \action_t \gets \policy_0(\state_t) \implies \\
			& \mquad{2} \forall \eps \ge \eps_0 \spc \exists \delta \ge 0 \spc \goaldist(\state_0) \le \delta \\
			& \mquad{3} \implies \forall t \ge 0 \spc \goaldist(\state_t) \le \eps, 
		\end{aligned}
	\end{equation}	
	where $\delta$ is unbounded as a function of $\eps$ and $\delta = 0 \iff \eps = \eps_0$.
	The presence of an $\eps_0$ means we do not insist on $\G$ being invariant.
	Thus, this extra condition only means that the state overshoot may be uniformly bounded over compacts in $\states$.
\end{dfn}

Notice that if the cost $\cost$ be continuous, strictly positive outside $\G$ and diverging with the distance to $\G$, the optimal policy $\policy^{\ast}$ is also a uniform $\G$-stabilizer.

The main hypothesis of this work is that incorporating some $\policy_0$ into the agent may improve its learning and guarantee stabilization into the goal set in all learning episodes \ie online.
The main problem is: how to do this incorporation?
Notice that even if $\policy_1, \policy_2$ are two stabilizers, an arbitrary switching or ``shaping" onto into another will not provide stabilization guarantee in general.

%%%% Some stuff on stabilizing policies
%
%such that the distance between $s_t$ and $\G \subset \states$ tends to zero if $\action_{t} = \policy(\state_{t})$ (a set of \textit{stabilizing policies}). Note that it does not necessarilly include all such policies;
%
%$\G$ is the \textit{target set}, that is a bounded neighbourhood of the origin in $\states$.

%%%%%%%%%%%%%%%%%%%%%%%%%%%%%%%%%%%%%%%%%%%%%%%%%%%%%%%%%%%%%%%%%%%%%%%%%%%%%%%%%%%%%%%%%%%%
%%%%%%%%%%%%%%%%%%%%%%%%%%%%%%%%%%%%%%%%%%%%%%%%%%%%%%%%%%%%%%%%%%%%%%%%%%%%%%%%%%%%%%%%%%%%
\section{Related work}
\label{sec_sota}

There are three principal methodologies for stability-ensuring reinforcement learning: shield-based reinforcement learning, the integration of MPC with reinforcement learning, and Lyapunov-based reinforcement learning.

Shield-based approaches involve a protective filter, also referred to as a shield, overseer, or supervisor, designed to prevent undesirable, say, destabilizing actions.
These methods vary in how they evaluate actions and generate safe alternatives.
Shields range from manual human oversight \cite{Saunders2017TrialerrorTow} to sophisticated formal verification variants \cite{Tan2020DeductiveStabi,Platzer2008Keymaerahybrid,Platzer2009EuropeanTrain,Fulton2018Safereinforcem}.
Unlike human operators, formal logic shields are theoretically error-proof, but they require highly specific application development and can be complex, as detailed in \cite{Koenighofer2020Shieldsynthesi}.
Such techniques have been applied in areas such as probabilistic shielding \cite{Koenighofer2020Safereinforcem,Koenighofer2020Shieldsynthesi}, supervisory systems for autonomous vehicles \cite{Isele2018Safereinforcem}, and safe robotic manipulation \cite{Thananjeyan2021RecoveryrlSaf}.
However, human overseers introduce subjective biases and potential errors, and formal logic shields are often difficult to design.

The combination of MPC and reinforcement learning represents an active frontier in the quest for ensuring stability \cite{Zanon2020SafeReinforcem,Zanon2019Practicalreinf,Koller2018LearningBased,Berkenkamp2017SafeModelbase,Berkenkamp2019Safeexploratio,Oh2023QMPCstable}.
Such a fusion takes various forms, sometimes emphasizing model learning and other times focusing on safety constraints \cite{Karnchanachari2020PracticalReinf,Lowrey2018Planonlinelea,Cai2023Energymanageme,Amos2018Differentiable,Hoeller2020DeepValueMode,East2020InfiniteHorizo,Reddy2019LearningHuman,Finn2017Deepvisualfor,Karnchanachari2020PracticalReinf,Asis2020FixedHorizonT}.
The reinforcement learning dreamer is a notable example that adopts the predictive spirit of MPC \cite{Hafner2020DreamControlL,Wu2023DaydreamerWorl}.
Such approaches draw on the MPC's well-established ability to ensure closed-loop stability via techniques like terminal costs and constraints.
Proposals such as the one by Zanon et al. \cite{Zanon2020SafeReinforcem,Zanon2019Practicalreinf} embed robust MPC within reinforcement learning to maintain safety.
Other research emphasizes the predictive control side more heavily \cite{Koller2018LearningBased,Berkenkamp2017SafeModelbase}, building from a safe starting point towards a predictive model while maintaining the option to revert to safe policies when needed.
%MPC integration with reinforcement learning ensures correctness through established safety and stability protocols, yet the intersection of these stability measures with learning presents areas ripe for further investigation, particularly in continuous-time and stochastic environments.

Lyapunov stability theory is well recognized in reinforcement learning, having roots dating back to the work of Perkins and Barto \cite{Perkins2002LyapunovDesign,Perkins2001Lyapunovconstr}, and has seen significant development since \cite{Chow2018Lyapunovbased,Berkenkamp2017SafeModelbase,Jeddi2023Memoryaugmente,Han2020Actorcriticre,Chang2021Stabilizingneu}.
Typically, these approaches are offline and require validation of a Lyapunov decay condition in the state space.
Chow et al. \cite{Chow2018Lyapunovbased}, for instance, developed a safe Bellman operator to ensure Lyapunov compliance, while Berkenkamp et al. \cite{Berkenkamp2017SafeModelbase} used state space segmentation to validate Lyapunov conditions, demanding certain confidence levels in the statistical environment model.
Online Lyapunov-based approaches also exist, often inspired by adaptive control techniques \cite{Zhang2011Datadrivenrob}.
Robustifying controls, a key component introduced by \cite{vamvoudakis2015asymptotically}, may distort the learning unless certain preconditions are met.
Some reviews may be found in \cite{Kamalapurkar2018Reinforcementl,Osinenko2022Reinforcementl}.
Control barrier functions, another safety feature, have been successfully integrated with reinforcement learning, providing enhanced safety capabilities as seen in simulations with bipedal robots \cite{Choi2020Reinforcementl} and in model-free reinforcement learning agents \cite{Cheng2019Endendsafe}.
Stochastic stability theory provides a basis for correctness \cite{Khasminskii2011StochasticStab}, but the current landscape of Lyapunov-based methods often lacks capacity for real-time application without extensive pre-training, and is generally predicated on specific assumptions about environmental dynamics, such as second-order differentiability \cite{Bhasin2013novelactorcri}, linearity \cite{Vamvoudakis2017Qlearningcont}, or global Lipschitz continuity \cite{Vrabie2012OptimalAdaptiv}.

It should be stressed that policy shaping algorithms \eg \cite{Plisnier2023TransferringMu} may be attractive in their similarity to pre-training the agent to boost learning, no online stabilization can be ensured under them.
In contrast to the existing approaches, CALF is online, bringing the interplay of a stabilizer and the agent onto a rigorous footing, thus providing a viable, model-free approach to combining classical control and reinforcement learning.
This is the essence of the current contribution.

%%%%%%%%%%%%%%%%%%%%%%%%%%%%%%%%%%%%%%%%%%%%%%%%%%%%%%%%%%%%%%%%%%%%%%%%%%%%%%%%%%%%%%%%%%%%
%%%%%%%%%%%%%%%%%%%%%%%%%%%%%%%%%%%%%%%%%%%%%%%%%%%%%%%%%%%%%%%%%%%%%%%%%%%%%%%%%%%%%%%%%%%%
\section{Approach}
\label{sec_approach}

In general, \blue{it} holds, due to the Hamilton-Jacobi-Bellman equation, that $\forall t \ge 0 \spc Q^*_{t+1} - Q^*_t = - \cost^*_t$, where $Q^*_t = \cost(\state_t, \policy^*(\state_t)) + \Value^{\policy^*} \left( \state^{\policy^*(\state_t)}_{t+1} \right), \cost^*_t = \cost(\state_t, \policy^*(\state_t))$.
This effectively means, given the conditions on the cost stated in Section \ref{sec_problem}, that $Q^*_t$ is a Lyapunov function for the closed-loop system $\state_{t+1} = \transit( \state_t, \policy^*(\state_t) )$.
If a model (critic) is employed \eg a deep neural network $\hat Q^w(\state, \action)$, due to imperfections of learning, the Lyapunov property may be lost, although desired.
In CALF, we would like to retain the Lyapunov property of $\hat Q^w$.
Enforcing Lyapunov-like constraints alone on $\hat Q^w$ would not solve the problem of ensuring stability, since those constraints may fail to be feasible at some time steps.
Offline approaches, as mentioned in Section \ref{sec_sota}, would overcome this by large samples of state-action trajectories.
What we do here instead is that we employ any stabilizer, let us call it $\policy_0$.
The latter may be synthesized by common control techniques, like PID, sliding mode or funnel control.
The question is, as stated earlier, how to combine the agent with $\policy_0$?
This is brought into a systemic way in CALF.
Namely, the critic update \ie update of the weights $w$ is done so as to try to satisfy Lyapunov-like constraints (decay and $\Kinf$-bounds, see Algorithm \ref{alg_calfq}, line 7).
If this succeeds, the weights are passed and the next actor update will base upon them.
If not, the weights are ``frozen" (the respective variable is denoted $w^\dagger$) and $\policy_0$ is invoked.
Along $w^\dagger$, the state-action pair is also ``stored'', namely, $\state^\dagger, \action^\dagger$.
The Lyapunov-like constraints are checked relative to $\state^\dagger, \action^\dagger, w^\dagger$.
This trick enables to safely combine the agent with $\policy_0$ so as to retain the stabilization guarantee of the latter.
We use the Q-function to make the overall approach model-free.
The actor loss $\loss^\act$ may be taken equal the current critic with last ``successful'' weights $w^\dagger$, namely, $\loss^\act(\action) = \hat Q^{w^\dagger}(\state_t, \action)$.
Any augmentation of the loss \eg penalties or entropy terms, may be included into  $\loss^\act$, there is no restriction.
Furthermore, one may choose to take an $\eps$-greedy action.
 
Regarding the $\Kinf$-bounds (see Algorithm \ref{alg_calfq}, line 7), a reasonable choice of $\hat \kappa_\low$, $\hat \kappa_\up$ would be
\begin{equation}
	\label{eqn_calfq_quadkapps}
	\hat \kappa_\low(\bullet) = C_\low \bullet^{2}, \ \hat \kappa_\up(\bullet) = C_\up \bullet^{2}, \spc 0 < C_\low < C_\up.
\end{equation}
For the decay constraint, one may also take a quadratic rate or simply a constant $\bar \nu > 0$.
Overall, the hyper-parameters $\hat \kappa_\low$, $\hat \kappa_\up$ and $\bar \nu$ determine a trade-off between freedom of learning and worst-case-scenario reaching time of the goal.
If $\frac{\hat \kappa_\low}{\hat \kappa_\up}$ and $\bar \nu$ are chosen to be sufficiently small, the weights of the critic will not be prevented from converging to their ideal values, however if the critic is underfitted, smaller values of $\frac{\hat \kappa_\low}{\hat \kappa_\up}$, $\bar \nu$ may entail slower stabilization accordingly.

Now, the critic loss $\loss^\crit$ (see Algorithm \ref{alg_calfq}, line 7) may be taken in various forms.
The presented approach does not restrict the user.
For instance, one may take a TD(1) on-policy loss as per:
\begin{equation}
	\label{eqn_TD1}
	\begin{aligned}
		\loss^\crit(w) = & \sum_{k = 0}^\Treplay \big( \hat Q^w(\state_{t_k}, \action_{t_k}) - \cost(\state_{t_k}, \action_{t_k}) - \\
		            & \quad \hat Q^{w^\dagger}( \state_{t_{k+1}}, \action_{t_{k+1}} ) \big)^2 + \lrcrit^{-2}\nrm{w - w^\dagger}^2.
	\end{aligned}
\end{equation}
The regularization term $\lrcrit^{-2}\nrm{w - w^\dagger}^2$ is redundant if gradient descent based minimization is used, since one could simply set a learning rate $\lrcrit$ as opposed to penalizing the displacement of weights.
Notice the choice of the critic loss (or learning rate) does not prevent environment stabilization, although the quality of the learning may be affected.
Finally, \eqref{eqn_TD1} is akin to the critic loss of SARSA due to its on-policy character, but this is not necessary, an off-policy loss may be used as well \eg with greedy actions instead of $\action_{t_{k+1}}$ in \eqref{eqn_TD1}.

%The stabilizing constraints read:
%\begin{equation}
%	\label{eqn_qstabconstr}
%	\begin{aligned}
%		 & \hat Q^w(\state_t, \action_t) - \hat Q^{w^\dagger}(\state^\dagger, \action^\dagger) \leq -{\bar \nu}, \\
%		 & \hat \kappa_\low (\nrm{\state_t}) \leq \hat Q^{w}(\state_t, \action_t) \le \hat \kappa_\up(\nrm{\state_t}).
%	\end{aligned}
%\end{equation}

\begin{algorithm}
	\begin{algorithmic}[1]
	\STATE \textbf{Input}: ${\bar \nu} > 0, \hat \kappa_\low, \hat \kappa_\up, \policy_0: \text{(uniform) stabilizer}$
	\STATE \textbf{Initialize}: $\state_0, \action_0 := \policy_0(\state_0), w_0$ \sut
	\[
		\hat \kappa_\low (\nrm{\state_0}) \le \hat Q^{w_0}(\state_0, \action_0) \le \hat \kappa_\up(\nrm{\state_0})
	\]
	\STATE $w^\dagger \gets w_{0}, \state^\dagger \gets \state_0, \action^\dagger \gets \policy_0(\state_0), \action_0 \gets \policy_0(\state_0)$
	\FOR {$t := 1, \dots \infty$}
		\STATE Take action $\action_{t -1}$, get state $\state_t$
	   	\STATE Update action: $\action^* \gets \argmin\limits_{\action \in \actions(\state_t)} \hat Q^{w^\dagger}(\state_t, \action)$
%	   	or \eg $\eps$-greedy 		
		\STATE Try critic update
%		\vspace*{-\baselineskip}
		\[
		\begin{array}{lll}
	 		w^*  \gets  &  \hspace{-5pt} \argmin\limits_{w \in \W} \loss^\crit(w) \\
				& \hspace{-8pt} \sut \; \hat Q^w(\state_t, \action_t) - \hat Q^{w^\dagger}(\state^\dagger, \action^\dagger) \le - {\bar \nu}, \\
				& \hspace{-8pt} \phantom{\sut \;} \hat \kappa_\low (\nrm{\state_t}) \le \hat Q^{w}(\state_t, \action_t) \le \hat \kappa_\up(\nrm{\state_t})
		\end{array}
		\]
%		\vspace*{-\baselineskip}
		\IF{ solution $w^*$ found} \alglinelabel{algline_calfcheck} 
		\STATE $\state^\dagger \gets \state_t, \action^\dagger \gets \action^*, w^\dagger \gets w^*$
		\ELSE
		\STATE $\action_t \gets \policy_0(\state_t)$	
		\ENDIF	
	\ENDFOR
	\end{algorithmic}
	\caption{Critic as Lyapunov function (CALF) algorithm, model-free, action-value based}
	\label{alg_calfq}
\end{algorithm}

\begin{rem}
\label{rem_actasmodel}
An actor model $\policy^\theta$ \eg as probability distribution, with weights $\theta$ may be employed instead of direct actions $a_t$ in Algorithm \ref{alg_calfq}.
\end{rem}

\subsection{Modified SARSA}
\label{sec_sarsam}

The new CALF agent was benchmarked via its immediate reinforcement learning alternative, \textit{State–action–reward–state–action} (SARSA), which is essentially like Algorithm \ref{alg_calfq} prescribes (with on-policy critic loss), but with the Lyapunov-like constraints and the $\policy_0$ removed.
In our case studies with a mobile robot, we observed such a plain SARSA failed to drive the robot to the target area within any reasonable number of learning iterations.
To help SARSA succeed, we slightly modified it, namely, we retained the $w^\dagger$-mechanism \ie we used $w^\dagger$ in the critic loss (see Algorithm \ref{alg_calfq}, line 7).
We did not \textit{enforce} the Lyapunov-like constraints in the optimization though.
We only checked, whether those constraints were satisfied post factum and updated $w^\dagger$ accordingly as in CALF.
Such a modification turned out to help SARSA reach the target in some learning runs.
This algorithm will further be referred to as SARSA-m.

%%%%%%%%%%%%%%%%%%%%%%%%%%%%%%%%%%%%%%%%%%%%%%%%%%%%%%%%%%%%%%%%%%%%%%%%%%%%%%%%%%%%%%%%%%%%
%%%%%%%%%%%%%%%%%%%%%%%%%%%%%%%%%%%%%%%%%%%%%%%%%%%%%%%%%%%%%%%%%%%%%%%%%%%%%%%%%%%%%%%%%%%%
\section{Analysis}
\label{sec_theorems}

The main environment stability result is formulated in Theorem \ref{thm_calfsstab}.

\begin{thm}
	\label{thm_calfsstab}
	Consider the problem \eqref{eqn_ocproblem} and Algorithm \ref{alg_calfq}.
	Let $\policy_t$ denote the policy generated by Algorithm \ref{alg_calfq}.
	If the policy $\policy_{0}$ is a stabilizer, then $\policy_t$ is a stabilizer.
	If the former is a uniform stabilizer, then so is $\policy_t$.
\end{thm}

\begin{prf}

	First, let us consider $\G'$ to be a closed superset of $\G$, where the Hausdorff distance between $\G$ and $\G'$ is non-zero.
	Let $h$ denote the said distance.
	
	Recalling Algorithm \ref{alg_calfq}, let us denote:
	\begin{equation}
	    \label{eqn_qdagger}
	    \begin{aligned}
	        & \hat Q^\dagger := \hat Q^{w^\dagger}(\state^\dagger, \action^\dagger).
	    \end{aligned}
	\end{equation}
	
	Next, we introduce:
	\begin{equation}
	    \label{calftimes}
	    \begin{aligned}
	        & \hat{\mathbb T} := \{t \in \Z_{\ge 0} : \text{successful critic update} \}, \\
	        & \mathbb T^{\policy_0} := \{0\} \cup \\
	        & \qquad \left\{t \in \N : \begin{array}{l}
	        \text{successful critic update at } t-1 \spc \land \\
	        \text{unsuccessful at } t 
	        \end{array}
	        \right\}.
	    \end{aligned}
	\end{equation}
	The former set represents the time steps at which the critic succeeds and the corresponding induced action will fire at the next step.
	The latter set consist of the time steps after each of which the critic first failed, discarding the subsequent failures if any occurred.
	
	Now, let us define:
	\begin{equation*}
	    %    \label{eqn_calf2notation1}
	    \begin{aligned}
	        & \hat Q^\dagger_t :=  \begin{cases}
	                            \hat Q_t, t \in \hat{\mathbb T},\\
	                            \hat Q^\dagger_{t-1}, \text{ otherwise}.
	                        \end{cases}
	    \end{aligned}
	\end{equation*}
%	and, similarly, $\hat Q^\dagger_t$ for ease of reference.
	
	Next, observe that there are at most
	\begin{equation}
	    \label{eqn_CALF2_critic_reaching}
	        \hat T := \max \left\{ \frac{ \hat Q^\dagger_0 - \bar \nu}{\bar \nu}, 0 \right\}
	\end{equation}
	steps until the critic stops succeeding and hence only $\policy_0$ is invoked from that moment on.
	Hence, $\hat{\mathbb T}$ is a finite set.
	Notice $\hat T$ was independent of $\G'$ and in turn dependent on the initial value of the critic. 
	
	Consider some $t^\dagger$, a time step after which the critic failed to find a solution.
	At step $t^\dagger+1$, the action $\action_{t^\dagger}$ is taken leading the state to transition into some $\state_{t^\dagger+1} = \transit(\state_{t^\dagger}, \action_{t^\dagger})$.
	Now, either the critic finds a solution $w^\dagger_{t^\dagger+1}$ again, or $\policy_0$ is invoked.
	
	In the latter case, by \eqref{eqn_goallim}, let $T^{\policy_0}_{t^\dagger+1}$ be \sut
	\[
		\forall t \ge t^\dagger+1 + T^{\policy_0}_{t^\dagger+1} \spc \goaldist(\state_t) \le h.
	\]
	In other words, $\G'$ would be reached in no more than $\forall t \ge T^{\policy_0}_{t^\dagger+1}$ steps.
	Notice $T^{\policy_0}_{t^\dagger+1}$ implicitly depends on $\state_{t^\dagger+1}$.
	
	Since there are at most $\hat T$ such episodes, where $\policy_0$ is invoked for at least one step, the set $\mathbb T^{\policy_0}$ is finite, although it implicitly depends on the initial state.
	
	Let $\bar T^{\policy_0}$ be the maximal number of steps among $T^{\policy_0}_{t^\dagger+1}, t^\dagger+1 \in \mathbb T^{\policy_0}$ \sut
	\[
		\forall t \ge t^\dagger+1 + T^{\policy_0}_{t^\dagger+1} \spc \goaldist(\state_t) \le h.
	\]
	That is, $\bar T^{\policy_0}$ is the longest time needed to reach $\G'$ starting from all, but finitely many states $\state_{t^\dagger+1}$ which are in turn uniquely determined by the initial state.
	
	We thus conclude that the set $\G'$ is reached in no more than
	\begin{equation}
		\label{eqn_totalreachtime}
	    T^* := \bar T^{\policy_0} \cdot \hat T
	\end{equation}
	steps.	
	Since $\G'$ was arbitrary, we conclude that $\policy_t$ is a stabilizer.
	Notice that the reaching time $T^*$ depends on the initial state $\state_0$ and cannot, in general, be made uniform over compacts in $\states$ where the environment starts.	
	
	Now, let us address the case where $\policy_t$ can indeed be a uniform stabilizer.
	For this, we need to demonstrate uniformity of state overshoot and, respectively, uniformity of $T^*$ on compacts in states.
	To this end, let $\states_0$ be any compact in $\states$ and define
	\begin{equation}
		\label{eqn_maxvel}
	    \bar \transit(\states_0) := \sup_{\state \in \states_0, \action \in \actions(\states_0))} \nrm{\transit(\state, \action)},
	\end{equation}
	which exists since $\transit$ is upper semi-continuous in norm and $\actions(\state)$ is compact for every state $\state$.

	Define
	\begin{equation*}
	  \begin{aligned}
	      & \state^\dagger_t :=  \begin{cases}
	                          \state_t, t \in \hat{\mathbb T},\\
	                          \state^\dagger_{t-1}, \text{ otherwise}.
	                      \end{cases}
	  \end{aligned}
	\end{equation*} 
	
	Observe that
	\[
  		\forall t \in \Z_{\ge 0} \spc \nrm{\state^\dagger_t} \le \hat \kappa_\low\inv(\hat Q^\dagger_t) \le \hat \kappa_\low\inv(\hat Q^\dagger_0).
	\]
	
	Denote
	\begin{equation}
		\label{eqn_initstatesdagger}
		\states_0^\dagger := \left\{ \state \in \states: \nrm{\state} \le \sup_{\state' \in \states_0} \hat \kappa_\low\inv( \hat \kappa_\up ( \nrm{\state'} ) ) \right\}.
	\end{equation}
	
	Denote the state trajectory induced by the policy $\policy_0$ emanating from $\state_0$ as $\traj_{0:\infty}^{\policy_0}(\state_0)$ with single elements thereof denoted $\traj_t^{\policy_0}(\state_0)$.
	By Proposition 2.2 in \cite{Jiang2002converseLyapun}, since $\policy_0$ is a uniform stabilizer, there exists a $\KL$ function $\beta$ \sut
	\begin{equation}
		\label{eqn_klstab}
		\forall t \in \Z_{\ge 0},  \state_0 \in \states \spc \goaldist( \traj_t^{\policy_0} ) \le \beta(\goaldist(\state_0), t).
	\end{equation}
	
	By \cite[Lemma~8]{Sontag1998Commentsintegr}, there exist two $\Kinf$ functions $\kappa, \xi$ \sut
	\begin{equation}
		\label{eqn_kldecompos}
		\forall v>0, t>0  \spc \beta(v, t) \le \kappa(v) \xi(e^{-t}).
	\end{equation}
	
	Since $0 \in \G$, it holds that
	\begin{equation}
		\label{eqn_kappagoal}
		\forall \state \in \states \spc \kappa( \goaldist(\state) ) \le \kappa(\nrm{\state}).
	\end{equation}
	
	Hence
	\begin{equation}
		\label{eqn_betabound2kappas}
		\forall \state \in \states, t>0  \spc \beta(\goaldist(\state), t) \le \kappa(\nrm{\state}) \xi(e^{-t}).
	\end{equation}		
	
	It holds that
	\begin{align*}
		& \forall t \in \Z_{\ge 0} \\
		& \quad \beta(\goaldist(\state^\dagger_t),0) \le \kappa(\goaldist(\state^\dagger_t)) \xi(1) \le \kappa( \hat \kappa_\low\inv(\hat Q^\dagger_t) ) \xi(1) \le \\
		& \mquad{4} \kappa( \hat \kappa_\low\inv(\hat Q^\dagger_0) ) \xi(1) \\
		& \pushright{ \le \kappa( \hat \kappa_\low\inv( \hat \kappa_\up( \nrm{\state_0} ) ) ) \xi(1). } 
	\end{align*}		
	
	Let us define
	\[
		\bar \beta^{\states_0^\dagger}:= \sup_{\nrm{\state} \le \bar \transit(\states_0^\dagger)} \beta(\state, 0).
	\]
	If $\policy_0$ were invoked from $t^\dagger+1$ on, the norm state would evolve bounded by $\beta(\goaldist(\state_{t+1}^\dagger), t-t^\dagger-1), t \ge t^\dagger+1$, due to \eqref{eqn_klstab}, until either $\G'$ would be reached or the critic would succeed again.
	This together with a maximal possible ``jump'' of $\beta$ lets us to deduce
	\[
		\forall t \spc \beta(\goaldist(\state_{t+1}^\dagger), 0) \le \beta(\goaldist(\state_{t}^\dagger), 0) + \bar \beta^{\states_0^\dagger}.
	\]
	 
	Now, let $\psi: \R \ra \R$ be defined as:
	\begin{equation}
		\label{eqn_globalcalfbound}
		\psi(v) := \kappa( \hat \kappa_\low\inv( \hat \kappa_\up( v ) ) ) \xi(1) + \bar \beta^{\states_0^\dagger} + \eps_0.
	\end{equation} 
	
	We deduce that
	\[
	    \begin{aligned}
	    & \forall t \in \Z_{\ge 0} \spc \goaldist(\state_t) \le \psi(\nrm{\state_0}).
	    \end{aligned}
	\]
	
	Hence, we may claim the state resides in a desired $\eps$-vicinity, with $\eps \ge \eps_0' := \bar \beta^{\states_0^\dagger} + \eps_0$, around $\G$ if the initial state satisfies
	\[
		\nrm{\state_0} \le \psi\inv\left( \eps \right).
	\]
	Notice the involved inverse exists due to the strict increase property.
	
	Let $\eps_\G$ be the Hausdorff distance between $\G$ and $\G'$.
	Now, we argue similarly to the non-uniform case as before, but defining
	\begin{equation*}
		\begin{aligned}
		    & \bar T^{\policy_0}_{\text{unif}} := \sup_{\nrm{\state} \le  \psi(\nrm{\state_0})  } \\
		    & \mquad{2} \max\left\{1, - \log \left( \xi\inv\left( \frac{\eps_\G}{\hat \kappa_\low\inv( \hat \kappa_\up( \nrm{\state}) )} \right) \right) \right\}.
	    \end{aligned}
	\end{equation*}	
	Notice this definition is now uniform over a compact.
	
	Notice also that it is straightforward to make $\hat T$ uniform over compacts in $\states$ \eg by setting
	\[
		\hat T := \max \left\{ \frac{\sup\limits_{\nrm{\state} \le  \psi(\nrm{\state_0}) } \hat \kappa_\up(\nrm{\state}) - \bar \nu}{\bar \nu}, 0 \right\}.
	\]
	The final reaching time reads $T^* = \hat T \cdot \bar T^{\policy_0}_{\text{unif}}$.
	
\end{prf}

\begin{rem}
	\label{rem_noovershoot}
	If $\G$ is contained in the set $\states_{\bar \nu} := \left\{ \state \in \states : \hat \kappa_\up( \nrm{\state}) \le \bar \nu  \right\}$, then $\eps_0' = \eps_0$, in particular, it is zero if $\G$ is invariant.
	This claim is true since if the state is in $\states_{\bar \nu}$, then the critic will not fire and hence only $\policy_0$ is invoked leading to the same overshoot as under $\policy_0$.
	It is verified by setting
	\begin{equation*}
		\psi(v) := \begin{cases}
			\kappa( \hat \kappa_\low\inv( \hat \kappa_\up( v ) ) \xi(1) + \bar \beta^{\states_0^\dagger} + \eps_0, v \ge \hat \kappa_\up\inv( \bar \nu ),  \\
			\kappa(v)\xi(1), \text{ otherwise}.
		\end{cases}
	\end{equation*}	
	Now, set $\bar \psi(v) := \sup_{v' \le v} \psi(v')$.
	It is easy to verify that $\bar \psi$ is an increasing function, hence it is Riemann integrable.
	Set 
	\[
		\Psi(v) := \int_{\eps_0}^{v} \bar \psi(v') \diff v'.
	\]	
	It is easy to verify that $\Psi$ is continuous, strictly increasing and bounds $\bar \psi$ from above.
	Hence, we may claim that the state resides in a desired $\eps$-vicinity, with $\eps \ge \eps_0$, around $\G$ if the initial state satisfies
	\[
		\nrm{\state_0} \le \Psi\inv\left( \eps \right).
	\]
	The same argumentation works if one sets in line \ref{algline_calfcheck} of Algorithm \ref{alg_calfq} the condition as ``\textbf{if} solution $w^*$ found and $\state_t \notin \hat \G$'' where $\hat \G$ is chosen to well contain $\G$.
	In that case, the critic deactivates in a vicinity of the goal.
	Notice otherwise Algorithm \ref{alg_calfq} does not rely on any explicit specification of the goal.
\end{rem}

\begin{rem}
	Since the number of invocations of $\policy_0$ is not greater than $T^*$ till the $\G$ is reached, the critic $\hat Q^\dagger$ is a multi-step Lyapunov function \ie $\hat Q^\dagger$ is non-increasing and
	\[
		\forall t \in \Z_{\ge 0} \spc \hat Q^\dagger_{t+T^*} - \hat Q^\dagger_t < 0.
	\]
\end{rem}

%\begin{rem}
%	The equation claim Theorem \ref{thm_calfsstab} can be rephrased in the following way: $\state_t$ will eventually get arbitrarily close to $\G$ and stay there permanently.
%	In other words, Theorem \ref{thm_calfsstab} states that Algorithm \ref{alg_calfq} computes a policy that is guaranteed to stabilize the environment.
%	This result is valuable, because it indicates that even a poorly trained reinforcement learning agent will still yield a stabilizing policy.
%\end{rem}

\begin{rem}
	For extensions of the stability result to stochastic environments and local basins of attraction, kindly refer to \cite{Osinenko2023actorcriticfr}.
\end{rem}

%%%%%%%%%%%%%%%%%%%%%%%%%%%%%%%%%%%%%%%%%%%%%%%%%%%%%%%%%%%%%%%%%%%%%%%%%%%%%%%%%%%%%%%%%%%%
%%%%%%%%%%%%%%%%%%%%%%%%%%%%%%%%%%%%%%%%%%%%%%%%%%%%%%%%%%%%%%%%%%%%%%%%%%%%%%%%%%%%%%%%%%%%
\section{Case study}
\label{sec_results}

\subsection{System description}

\begin{figure}
	\centering
	\includegraphics{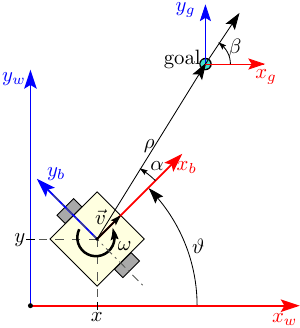}
	\caption{Robot kinematics and its frames of interests.}
	\label{fig_3wrobot}
\end{figure}

We consider here a differential drive mobile wheeled robot (WMR) depicted in Fig. \ref{fig_3wrobot} for testing the proposed algorithms.
The goal is to achieve autonomous stabilization of the robot at the origin of a coordinate system, starting from non-zero initial conditions.
%This means the robot must move from an arbitrary starting position to the origin (0,0) on a 2D plane.
In the work space, there is a designated high-cost zone, which could represent a physical object like a swamp, a puddle, or a phenomenon that can adversely impact the robot's movement. 
Within this high-cost zone, the robot's maximum velocity is constrained to 1 cm per second to mitigate potential damage or difficulties in navigation.
	
The differential drive model describes the motion of the wheeled robot.
In this model, the robot has two wheels mounted on either side, which can rotate independently. 
The position and orientation of the robot in the plane can be described by the state variables \((x, y, \vartheta)\), where:
	
\begin{itemize}
	\item \(x\) and \(y\) represent the coordinates of the robot in the plane.
	\item \(\vartheta\) is the orientation angle of the robot with respect to a reference direction.
\end{itemize}
	
The kinematic equations governing the motion of the robot are:
	
\begin{equation}
	\begin{aligned}
		\dot{x} &= v \cos(\vartheta) \\
		\dot{y} &= v \sin(\vartheta) \\
		\dot{\vartheta} &= \omega
	\end{aligned}
\end{equation}
	
where:
\begin{itemize}
	\item \(v\) is the linear velocity of the robot.
	\item \(\omega\) is the angular velocity (rate of change of orientation).
\end{itemize}
		
In the high-cost zone, the linear velocity $v$ is restricted such that $v \leq 1 \text{ cm/s}$. 
Outside of this zone, the robot can move at its maximum possible speed.

The control objective is to design a control law for $v$ and $\omega$ to stabilize the robot at the origin $(0,0)$. 
This involves navigating from any initial position to the origin while possibly avoiding the high-cost zone.
Experiments were conducted to validate the proposed algorithms under various initial conditions.
The performance metrics typically include the time taken to reach the origin, the path taken by the robot, and its behavior in and around the high-cost zone.

The performance of CALF and SARSA-m as the benchmark agent were compared with a nominal stabilizer, and two MPC agents with different horizons.
The latter are taken to get an idea about a nearly optimal controller for the stated problem.
The code for the environment simulation, CALF, SARSA-m, MPC and nominal stabilizer may be found under \url{https://github.com/osinenkop/rcognita-calf}.

In our studies, we observed that a significant horizon of length 15 was sufficient to fulfill the stated goal.
This controller is further referred to as MPC15.
Notice that MPC15 may lack practicability due to such a long horizon, especially taking into account possible disturbances.
Essentially, as the horizon length increases, the number of decision variables and constraints in the optimization problem grows, leading to increased computational complexity. 
Solving this problem within a reasonable time frame becomes more challenging as the horizon grows.
For real-time applications, such as controlling a differential drive mobile robot, the control decisions need to be computed and applied within very short time intervals. 
The extensive computations required for a 15-step horizon can exceed the available computational resources and time, making it possibly impractical for real-time implementation.
In this regard, CALF and SARSA-m are beneficial as they do not use any models.
Again, MPC15 was rather taken to get a clue of the best possible cost.

%Now, we get to the optimal control problem.
%The base environment is a wheeled mobile robot (WMR) depicted in Fig. \ref{fig_3wrobot}.
Recapping on the system description, at time $t$, the origin of the local coordinate frame (base) of the robot is located at the point $(x_t, y_t)$ of the world coordinate frame. 
The robot's linear velocity $v(t)$ along the axes determines the direction of motion, while the angular velocity $\omega(t)$ determines the rotation (refer to Fig.~\ref{fig_3wrobot}).
% Thus, the input of the system is a vector of two components $u(t) = \begin{bmatrix} v(t) & \omega(t) \end{bmatrix}$.
The state vector $\state \in \mathbb{R}^3$ and the control input vector (action) $\action \in \mathbb{R}^2$ are defined as

\begin{equation}
	\label{eqn_state_action_3wrobot}
	s \coloneqq \begin{bmatrix} x & y & \vartheta \end{bmatrix}^\top, \action \coloneqq \begin{bmatrix} v & \omega \end{bmatrix}^\top.
\end{equation}

The environment state transition map $\transit$ can be obtained via time discretization of the WMR differential equations which read: $\dot x = v \cos \vartheta, \dot y = v \sin \vartheta, \dot \vartheta = \omega$.
For the studied MPC, we used Euler discretization, whereas for CALF and SARSA-m, $\transit$ is not necessary at all.
In our case study, we used the controller sampling time of 100 ms.

The control actions were constrained by $\action_\min$ and $\action_\max$, which are determined by the actuators' capabilities.
We took those constraints matching Robotis TurtleBot3, namely, normally (except for the high-cost spot) 0.22 m/s of maximal magnitude of translational and 2.84 rad/s of rotational velocity, respectively.

The control goal is to drive the robot into $\G = \{ \state \in \R^3: \nrm{\state - \state^*} \le \Delta \}$, where the target state $\state^*$ was taken identical with the origin and $\Delta$ is target pose tolerance.

Regarding the high-cost zone, we introduced a spot on the plane $R^2$, namely, the cost was defined as (we did not penalize the action and only slightly penalized the angle):
\begin{equation}
	\label{eqn_3wrobotcost}
	\cost \left(\state, \action \right) = x^2 + y^2 + 0.1 \vartheta^2 + 10 \cost' \left(x, y \right), 
\end{equation}
where 
\begin{equation}
	\label{eqn_hotspot}
%	\begin{aligned}
	\cost'(x, y) = \frac{1}{2 \pi \sigma_x \sigma_y } \exp \Bigg(- \frac{1}{2} \left( \frac{\Delta x ^2}{\sigma_x ^ 2}  + \frac{\Delta y ^2}{\sigma_y ^ 2} \right) \Bigg),
%	\end{aligned}
\end{equation}
with $\Delta x = x - \mu_x, \Delta_y = y - \mu_y$; $\sigma_x = \sigma_y = 0.1$ being standard deviations in the along-track and cross-track directions, respectively; and $\mu_x = -0.6, \mu_y = -0.5$ being the high cost spot center. 

We now proceed to the description of the nominal stabilizer.
The action components $v, \omega$ were determined based on the polar coordinate representation of the WMR as per \cite{Astolfi1995Exponentialsta}, namely:
\begin{equation}
	\begin{bmatrix}
		\dot{\rho} \\
		\dot{\alpha} \\
		\dot{\beta}
	\end{bmatrix} = \begin{bmatrix}
		\pm \cos {\alpha} & 0 \\
		\mp \frac{\sin{\alpha}}{\rho} & 1 \\
		\pm \frac{\sin{\alpha}}{\rho} & 0
	\end{bmatrix}     \begin{bmatrix}
		\pm v \\
		\omega
	\end{bmatrix},
\end{equation}
where the top sign (plus or minus) holds if $\alpha \in (- \frac \pi 2, \frac \pi 2]$ and the bottom sign holds if $\alpha \in \left(-\pi, -\pi/2 \right]\cup \left(\pi/2, \pi \right]$.
The transformation into the polar coordinates in turn reads:
\begin{align*}
	\rho &= \sqrt{x ^ 2 + y ^ 2},\\
	\alpha &= -\vartheta + \arctan2(y, x),\\
	\beta &= -\vartheta - \alpha.
	\label{eqn_polar}
\end{align*} 

The stabilizer was set as per:
\begin{align}
	v & \la K_{\rho} \rho,\\
	\omega & \la K_{\alpha}\alpha + K_{\beta} \beta,
\end{align}
where $K_{\rho} > 0, K_{\beta} < 0, K_{\alpha} - K_{\rho} > 0$.

\begin{figure*}
	\centering
	\begin{subfigure}[t]{0.46\textwidth}
		\includegraphics[width=\linewidth]{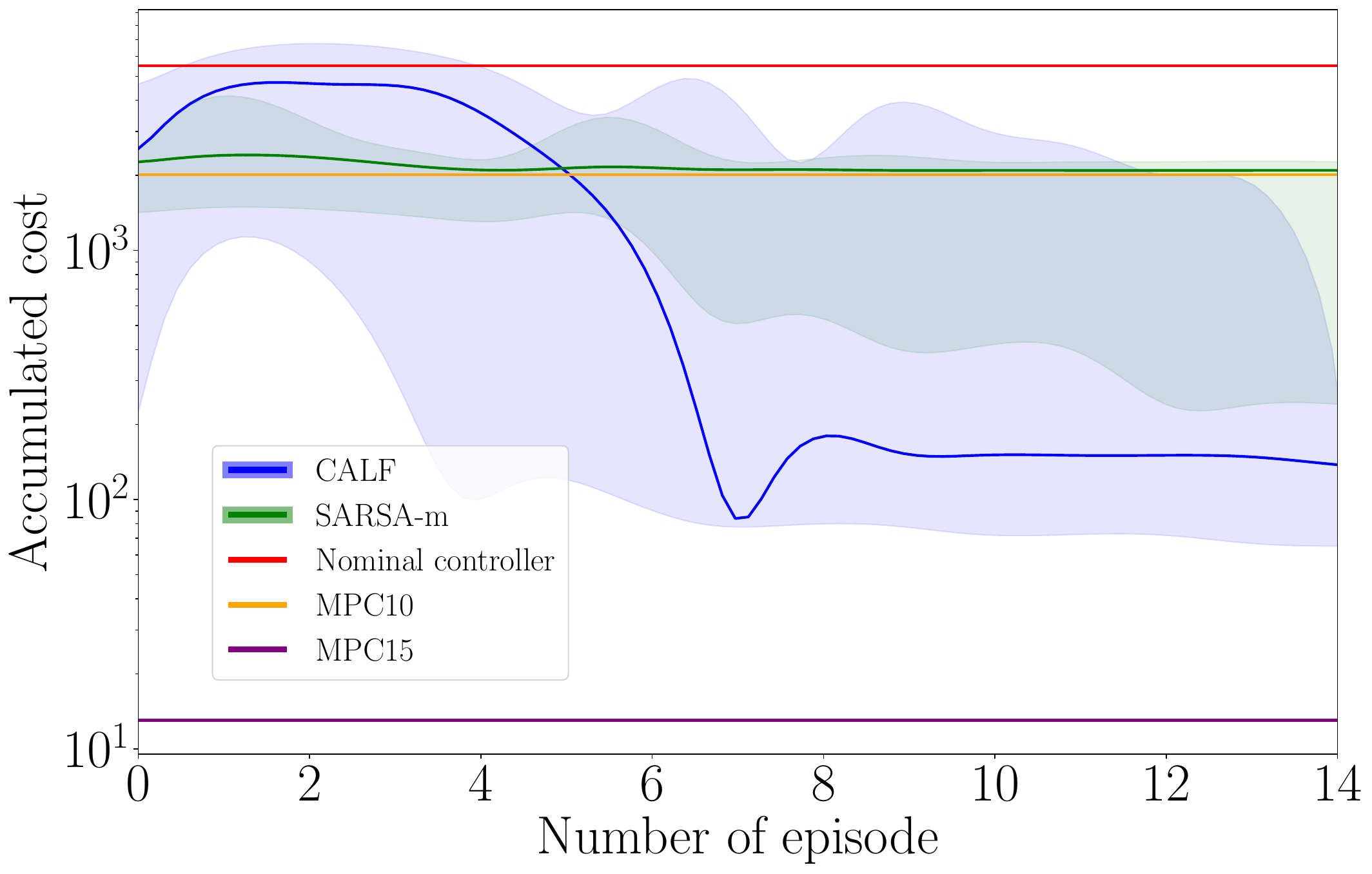}
		\caption{Learning curves depending on the episode \ie environment run number. The clouds represent the 95 \% confidence intervals. The solid lines are median. Model-predictive controllers, as well as the nominal stabilizer, are given for reference.}
		\label{fig_lrcurves}
	\end{subfigure}
	\phantom{---}
	\begin{subfigure}[t]{0.45\textwidth}
		\includegraphics[width=\linewidth]{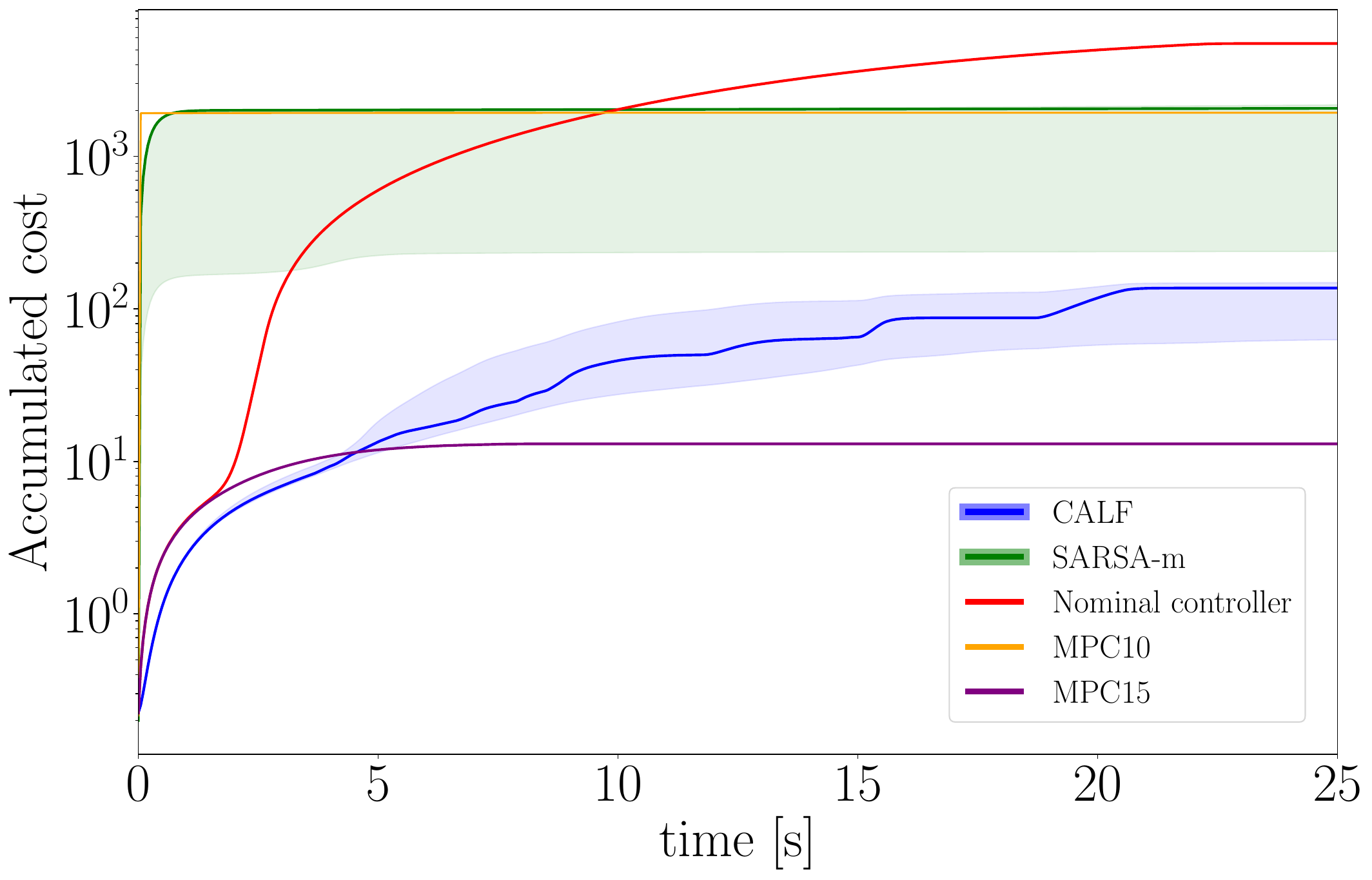}
		\caption{Accumulated cost (logarithmic scale) depending on time in the best learning episode. The clouds represent the 95 \% confidence intervals. The solid lines are medians.}
		\label{fig_accumcost}
	\end{subfigure}
	\caption{Learning curves obtained from 25 seeds of random number generator.}
	\label{fig_cost}
\end{figure*}

\begin{figure*}
	\centering
%			\begin{subfigure}[t]{0.32\textwidth}
%				\includegraphics[width=\linewidth]{gfx/traj1.pdf}
%				\caption{The final trajectories provided by the best learning models (according to the final cost metric).}
%				\label{fig:traj1}
%			\end{subfigure}
	\begin{subfigure}[t]{0.45\textwidth}
		\includegraphics[width=\linewidth]{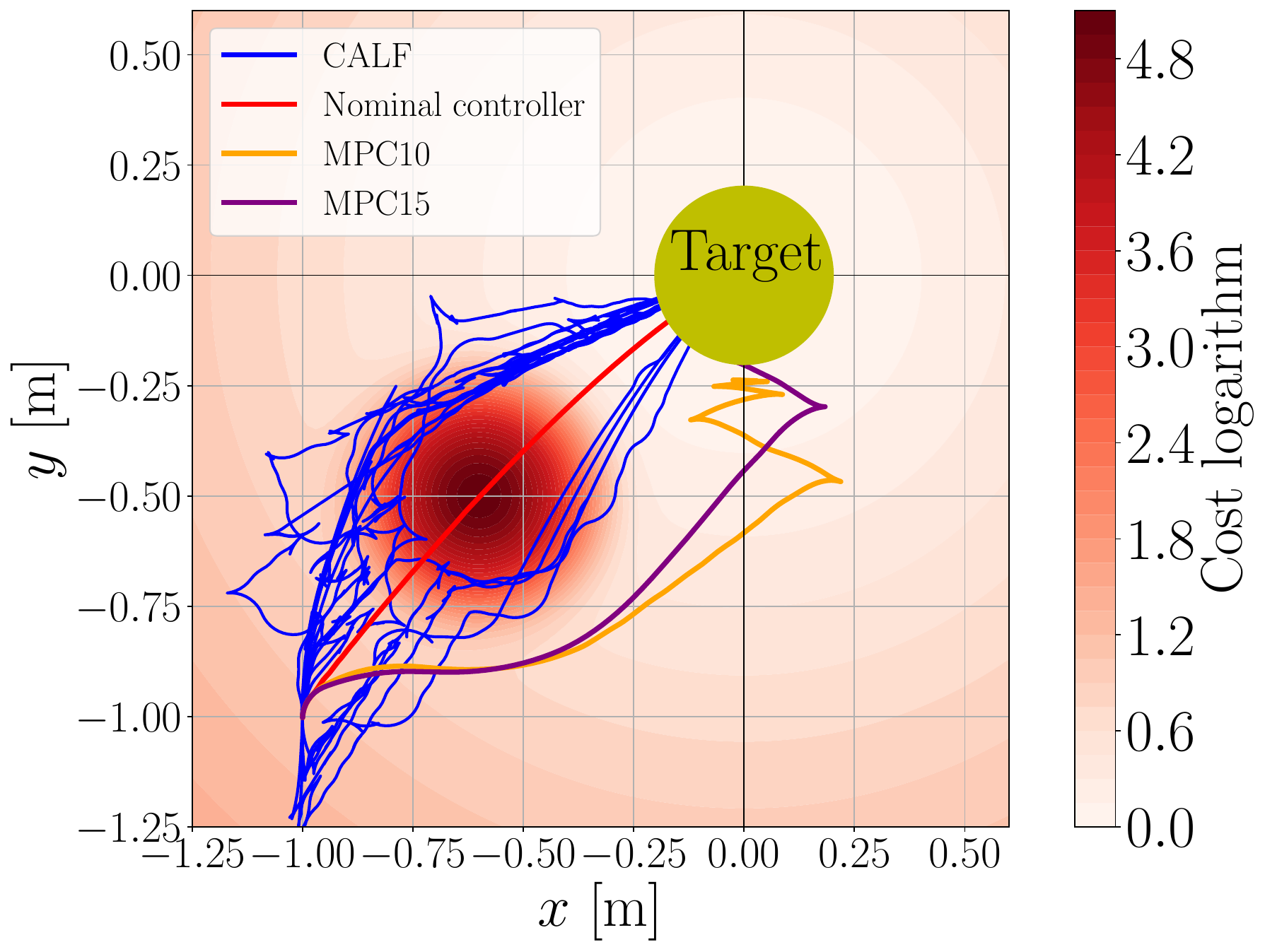}
		\label{fig:traj2}
	\end{subfigure}
	\phantom{---}
	\begin{subfigure}[t]{0.465\textwidth}
		\includegraphics[width=\linewidth]{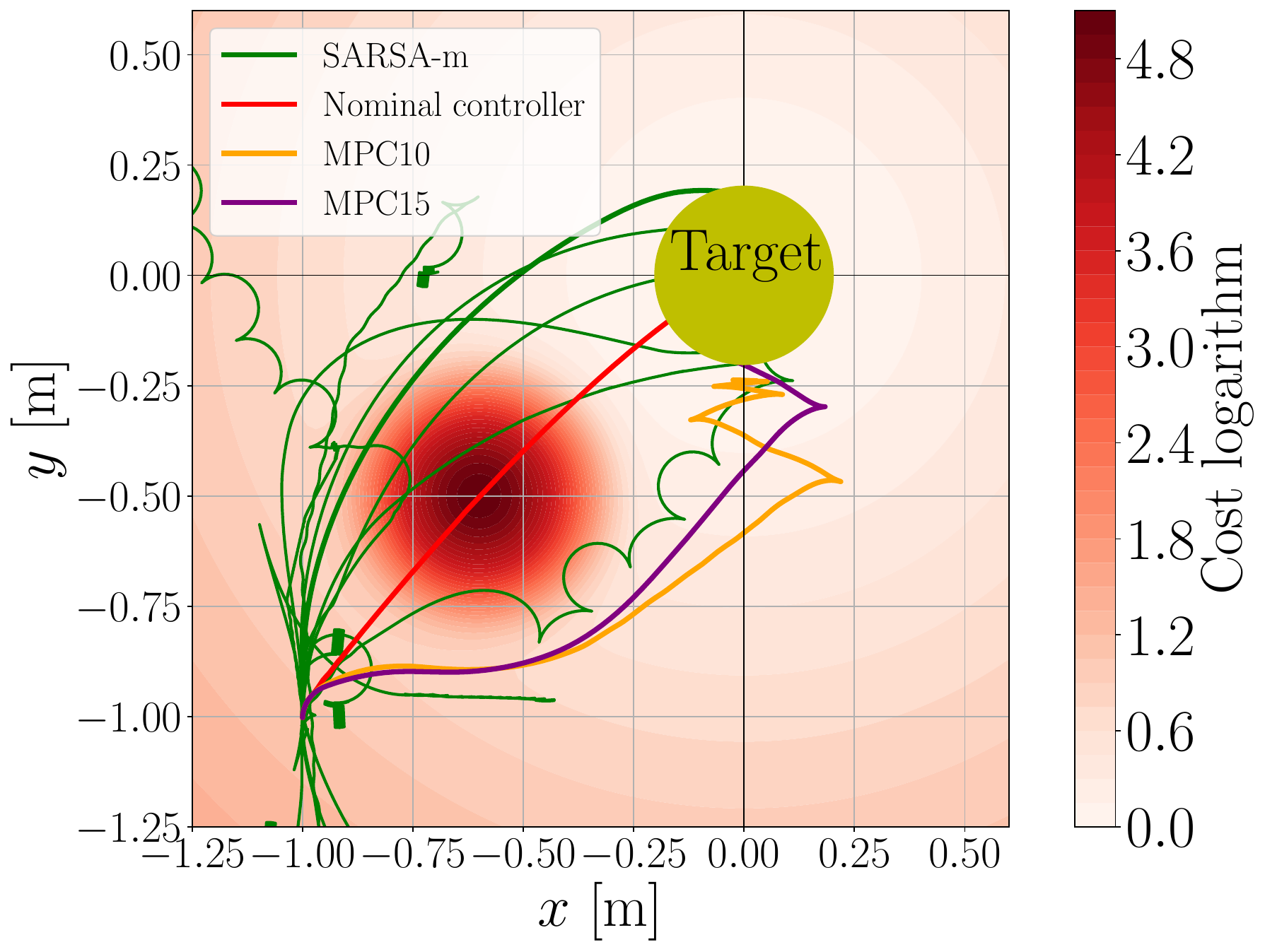}
		\label{fig:traj3}
	\end{subfigure}
\caption{The robot trajectories in best learning episodes over 25 seeds.}
\label{fig_traj}
\end{figure*}

The robot was run for 30 seconds in each episode.
If the target was not reached, we nominally added a cost of 2000 to rule out the cases where the robot simply did not drive to the target while also not getting into the ``hot'' spot determined by $\cost'$.
The target area was to 0.2 in radius around the origin (see Fig.~\ref{fig_traj}).
The initial robot pose was set to $\begin{Bmatrix}  -1 & -1 & \pi/2	\end{Bmatrix}$.

%%%%%%%%%%%%%%%%%%%%%%%%%%%%%%%%%%%%%%%%%%%%%%%%%%%%%%%%%%%%%%%%%%%%%%%%%%%%%%%%%%%%%%%%%%%%
\subsection{Discussion of results and conclusion}

As can be seen from the results (Fig.~\ref{fig_cost}, Fig.~\ref{fig_traj}), both reinforcement learning  agents outperformed MPC10 (CALF at once and SARSA-m in the final episode).
The controller MPC15 performed best in terms of cost, but it should only be taken as a reference for a potential nearly best performing approach.
It has a significant horizon length which reduces its practicability.
Unlike MPC15, the studied reinforcement learning agents are totally model-free.
Next, CALF also outperformed the nominal controller.
What is remarkable is that both agents were able to detour the ``hot'' spot (notice in Fig.~\ref{fig_traj} how the nominal stabilizer is blind to the spot).
CALF did it in all best learning episodes, whereas SARSA-m succeeded only in a part of those.
It should be repeated here that plain SARSA did not succeed in our case studies at all.
Overall, CALF always succeeded in reaching the target, thanks to its design and Theorem \ref{thm_calfsstab}, and also outperformed the benchmark agent SARSA-m in terms of learning performance.
This validates the initial claim that ensuring online environment stability is not just practicable, but it is also beneficial for episode-to-episode learning.

%%%%%%%%%%%%%%%%%%%%%%%%%%%%%%%%%%%%%%%%%%%%%%%%%%%%%%%%%%%%%%%%%%%%%%%%%%%%%%%%%%%%%%%%%%%%
%%%%%%%%%%%%%%%%%%%%%%%%%%%%%%%%%%%%%%%%%%%%%%%%%%%%%%% BIBLIOGRAPHY

\bibliographystyle{IEEEtran}
\bibliography{
bib/Osinenko__Dec2023,
bib/AIDA__Mar2024
}

% Generated by IEEEtran.bst, version: 1.14 (2015/08/26)
\begin{thebibliography}{10}
\providecommand{\url}[1]{#1}
\csname url@samestyle\endcsname
\providecommand{\newblock}{\relax}
\providecommand{\bibinfo}[2]{#2}
\providecommand{\BIBentrySTDinterwordspacing}{\spaceskip=0pt\relax}
\providecommand{\BIBentryALTinterwordstretchfactor}{4}
\providecommand{\BIBentryALTinterwordspacing}{\spaceskip=\fontdimen2\font plus
\BIBentryALTinterwordstretchfactor\fontdimen3\font minus
  \fontdimen4\font\relax}
\providecommand{\BIBforeignlanguage}[2]{{%
\expandafter\ifx\csname l@#1\endcsname\relax
\typeout{** WARNING: IEEEtran.bst: No hyphenation pattern has been}%
\typeout{** loaded for the language `#1'. Using the pattern for}%
\typeout{** the default language instead.}%
\else
\language=\csname l@#1\endcsname
\fi
#2}}
\providecommand{\BIBdecl}{\relax}
\BIBdecl

\bibitem{Sutton2018ReinforcementL}
R.~S. Sutton and A.~G. Barto, \emph{Reinforcement Learning: An
  Introduction}.\hskip 1em plus 0.5em minus 0.4em\relax Cambridge, MA, USA: A
  Bradford Book, 2018.

\bibitem{Kakade2001naturalpolicy}
S.~M. Kakade, ``A natural policy gradient,'' \emph{Advances in neural
  information processing systems}, vol.~14, 2001.

\bibitem{Baxter2001Infinitehorizo}
J.~Baxter and P.~L. Bartlett, ``Infinite-horizon policy-gradient estimation,''
  \emph{Journal of Artificial Intelligence Research}, vol.~15, pp. 319--350,
  2001.

\bibitem{Peters2006Policygradient}
J.~Peters and S.~Schaal, ``Policy gradient methods for robotics,'' in
  \emph{2006 IEEE/RSJ International Conference on Intelligent Robots and
  Systems}, 2006, pp. 2219--2225.

\bibitem{Bertsekas2019Reinforcementl}
D.~P. Bertsekas, \emph{Reinforcement learning and optimal control}.\hskip 1em
  plus 0.5em minus 0.4em\relax Athena Scientific Belmont, MA, 2019.

\bibitem{Sutton1991Dynaintegrated}
R.~S. Sutton, ``Dyna, an integrated architecture for learning, planning, and
  reacting,'' \emph{ACM Sigart Bulletin}, vol.~2, no.~4, pp. 160--163, 1991.

\bibitem{Pei2021improveddynaq}
M.~Pei, H.~An, B.~Liu, and C.~Wang, ``An improved dyna-q algorithm for mobile
  robot path planning in unknown dynamic environment,'' \emph{IEEE Transactions
  on Systems, Man, and Cybernetics: Systems}, vol.~52, no.~7, pp. 4415--4425,
  2021.

\bibitem{Saunders2017TrialerrorTow}
W.~Saunders, G.~Sastry, A.~Stuhlmueller, and O.~Evans, ``Trial without error:
  Towards safe reinforcement learning via human intervention,'' \emph{arXiv
  preprint arXiv:1707.05173}, 2017.

\bibitem{Tan2020DeductiveStabi}
Y.~K. Tan and A.~Platzer, ``Deductive stability proofs for ordinary
  differential equations,'' \emph{arXiv:2010.13096}, 2020.

\bibitem{Platzer2008Keymaerahybrid}
A.~Platzer and J.-D. Quesel, ``Keymaera: A hybrid theorem prover for hybrid
  systems (system description),'' in \emph{Automated Reasoning}.\hskip 1em plus
  0.5em minus 0.4em\relax Springer, 2008, pp. 171--178.

\bibitem{Platzer2009EuropeanTrain}
------, ``European train control system: A case study in formal verification,''
  in \emph{Formal Engineering Methods}.\hskip 1em plus 0.5em minus 0.4em\relax
  Springer, 2009, pp. 246--265.

\bibitem{Fulton2018Safereinforcem}
N.~Fulton and A.~Platzer, ``Safe reinforcement learning via formal methods:
  Toward safe control through proof and learning,'' in \emph{Proceedings of the
  AAAI Conference on Artificial Intelligence}, vol.~32, no.~1, 2018.

\bibitem{Koenighofer2020Shieldsynthesi}
B.~K{\"o}nighofer, F.~Lorber, N.~Jansen, and R.~Bloem, ``Shield synthesis for
  reinforcement learning,'' in \emph{International Symposium on Leveraging
  Applications of Formal Methods}.\hskip 1em plus 0.5em minus 0.4em\relax
  Springer, 2020, pp. 290--306.

\bibitem{Koenighofer2020Safereinforcem}
B.~K{\"o}nighofer, R.~Bloem, S.~Junges, N.~Jansen, and A.~Serban, ``Safe
  reinforcement learning using probabilistic shields,'' in \emph{International
  Conference on Concurrency Theory: 31st CONCUR 2020: Vienna, Austria (Virtual
  Conference)}.\hskip 1em plus 0.5em minus 0.4em\relax Schloss
  Dagstuhl-Leibniz-Zentrum fur Informatik GmbH, Dagstuhl Publishing, 2020.

\bibitem{Isele2018Safereinforcem}
D.~Isele, A.~Nakhaei, and K.~Fujimura, ``Safe reinforcement learning on
  autonomous vehicles,'' in \emph{2018 IEEE/RSJ International Conference on
  Intelligent Robots and Systems (IROS)}, 2018, pp. 1--6.

\bibitem{Thananjeyan2021RecoveryrlSaf}
B.~Thananjeyan, A.~Balakrishna, S.~Nair, M.~Luo, K.~Srinivasan, M.~Hwang, J.~E.
  Gonzalez, J.~Ibarz, C.~Finn, and K.~Goldberg, ``Recovery rl: Safe
  reinforcement learning with learned recovery zones,'' \emph{IEEE Robotics and
  Automation Letters}, vol.~6, no.~3, pp. 4915--4922, 2021.

\bibitem{Zanon2020SafeReinforcem}
M.~Zanon and S.~Gros, ``Safe reinforcement learning using robust {MPC},''
  \emph{{IEEE} Transactions on Automatic Control}, vol.~66, no.~8, pp.
  3638--3652, 2020.

\bibitem{Zanon2019Practicalreinf}
M.~Zanon, S.~Gros, and A.~Bemporad, ``Practical reinforcement learning of
  stabilizing economic mpc,'' in \emph{2019 18th European Control Conference
  (ECC)}, 2019, pp. 2258--2263.

\bibitem{Koller2018LearningBased}
T.~Koller, F.~Berkenkamp, M.~Turchetta, and A.~Krause, ``Learning-based model
  predictive control for safe exploration,'' in \emph{2018 {IEEE} Conference on
  Decision and Control ({CDC})}.\hskip 1em plus 0.5em minus 0.4em\relax {IEEE},
  dec 2018.

\bibitem{Berkenkamp2017SafeModelbase}
F.~Berkenkamp, M.~Turchetta, A.~Schoellig, and A.~Krause, ``Safe model-based
  reinforcement learning with stability guarantees,'' in \emph{Advances in
  Neural Information Processing Systems}, I.~Guyon, U.~V. Luxburg, S.~Bengio,
  H.~Wallach, R.~Fergus, S.~Vishwanathan, and R.~Garnett, Eds., vol.~30.\hskip
  1em plus 0.5em minus 0.4em\relax Curran Associates, Inc., 2017.

\bibitem{Berkenkamp2019Safeexploratio}
F.~Berkenkamp, ``Safe exploration in reinforcement learning: Theory and
  applications in robotics,'' Ph.D. dissertation, ETH Zurich, 2019.

\bibitem{Oh2023QMPCstable}
T.~H. Oh, ``Q-mpc: stable and efficient reinforcement learning using model
  predictive control,'' \emph{IFAC-PapersOnLine}, vol.~56, no.~2, pp.
  2727--2732, 2023.

\bibitem{Karnchanachari2020PracticalReinf}
N.~Karnchanachari, M.~de~la Iglesia~Valls, D.~Hoeller, and M.~Hutter,
  ``Practical reinforcement learning for mpc: Learning from sparse objectives
  in under an hour on a real robot,'' in \emph{Proceedings of the 2nd
  Conference on Learning for Dynamics and Control}, ser. Proceedings of Machine
  Learning Research, A.~M. Bayen, A.~Jadbabaie, G.~Pappas, P.~A. Parrilo,
  B.~Recht, C.~Tomlin, and M.~Zeilinger, Eds., vol. 120.\hskip 1em plus 0.5em
  minus 0.4em\relax The Cloud: PMLR, 2020, pp. 211--224.

\bibitem{Lowrey2018Planonlinelea}
K.~Lowrey, A.~Rajeswaran, S.~Kakade, E.~Todorov, and I.~Mordatch, ``Plan
  online, learn offline: Efficient learning and exploration via model-based
  control,'' \emph{arXiv preprint arXiv:1811.01848}, 2018.

\bibitem{Cai2023Energymanageme}
W.~Cai, A.~B. Kordabad, and S.~Gros, ``Energy management in residential
  microgrid using model predictive control-based reinforcement learning and
  shapley value,'' \emph{Engineering Applications of Artificial Intelligence},
  vol. 119, p. 105793, 2023.

\bibitem{Amos2018Differentiable}
B.~Amos, I.~Jimenez, J.~Sacks, B.~Boots, and J.~Z. Kolter, ``Differentiable
  {MPC} for end-to-end planning and control,'' in \emph{Advances in Neural
  Information Processing Systems}, S.~Bengio, H.~Wallach, H.~Larochelle,
  K.~Grauman, N.~Cesa-Bianchi, and R.~Garnett, Eds., vol.~31.\hskip 1em plus
  0.5em minus 0.4em\relax Curran Associates, Inc., 2018, pp. 8289--8300.

\bibitem{Hoeller2020DeepValueMode}
D.~Hoeller, F.~Farshidian, and M.~Hutter, ``Deep value model predictive
  control,'' in \emph{Proceedings of the Conference on Robot Learning}, ser.
  Proceedings of Machine Learning Research, L.~P. Kaelbling, D.~Kragic, and
  K.~Sugiura, Eds., vol. 100, 2020, pp. 990--1004.

\bibitem{East2020InfiniteHorizo}
S.~East, M.~Gallieri, J.~Masci, J.~Koutnik, and M.~Cannon, ``Infinite-horizon
  differentiable model predictive control,'' \emph{International Conference on
  Learning Representations}, Jan. 2020.

\bibitem{Reddy2019LearningHuman}
S.~Reddy, A.~D. Dragan, S.~Levine, S.~Legg, and J.~Leike, ``Learning human
  objectives by evaluating hypothetical behavior,'' \emph{International
  Conference on Machine Learning}, 2019.

\bibitem{Finn2017Deepvisualfor}
C.~Finn and S.~Levine, ``Deep visual foresight for planning robot motion,'' in
  \emph{2017 IEEE International Conference on Robotics and Automation (ICRA)},
  2017, pp. 2786--2793.

\bibitem{Asis2020FixedHorizonT}
K.~D. Asis, A.~Chan, S.~Pitis, R.~Sutton, and D.~Graves, ``Fixed-horizon
  temporal difference methods for stable reinforcement learning,''
  \emph{Proceedings of the {AAAI} Conference on Artificial Intelligence},
  vol.~34, no.~04, pp. 3741--3748, 2020.

\bibitem{Hafner2020DreamControlL}
D.~Hafner, T.~Lillicrap, J.~Ba, and M.~Norouzi, ``Dream to control: Learning
  behaviors by latent imagination,'' \emph{International Conference on Learning
  Representations}, 2020.

\bibitem{Wu2023DaydreamerWorl}
P.~Wu, A.~Escontrela, D.~Hafner, P.~Abbeel, and K.~Goldberg, ``Daydreamer:
  World models for physical robot learning,'' in \emph{Conference on Robot
  Learning}.\hskip 1em plus 0.5em minus 0.4em\relax PMLR, 2023, pp. 2226--2240.

\bibitem{Perkins2002LyapunovDesign}
T.~Perkins and A.~Barto, ``Lyapunov design for safe reinforcement learning
  control,'' in \emph{Safe Learning Agents: Papers from the 2002 AAAI
  Symposium}, 2002, pp. 23--30.

\bibitem{Perkins2001Lyapunovconstr}
T.~J. Perkins and A.~G. Barto, ``Lyapunov-constrained action sets for
  reinforcement learning,'' in \emph{ICML}, vol.~1, 2001, pp. 409--416.

\bibitem{Chow2018Lyapunovbased}
Y.~Chow, O.~Nachum, E.~Duenez-Guzman, and M.~Ghavamzadeh, ``A lyapunov-based
  approach to safe reinforcement learning,'' in \emph{Advances in Neural
  Information Processing Systems}, S.~Bengio, H.~Wallach, H.~Larochelle,
  K.~Grauman, N.~Cesa-Bianchi, and R.~Garnett, Eds., vol.~31.\hskip 1em plus
  0.5em minus 0.4em\relax Curran Associates, Inc., 2018.

\bibitem{Jeddi2023Memoryaugmente}
A.~B. Jeddi, N.~L. Dehghani, and A.~Shafieezadeh, ``Memory-augmented
  lyapunov-based safe reinforcement learning: end-to-end safety under
  uncertainty,'' \emph{IEEE Transactions on Artificial Intelligence}, 2023.

\bibitem{Han2020Actorcriticre}
M.~Han, L.~Zhang, J.~Wang, and W.~Pan, ``Actor-critic reinforcement learning
  for control with stability guarantee,'' \emph{IEEE Robotics and Automation
  Letters}, vol.~5, no.~4, pp. 6217--6224, 2020.

\bibitem{Chang2021Stabilizingneu}
Y.-C. Chang and S.~Gao, ``Stabilizing neural control using self-learned almost
  lyapunov critics,'' in \emph{2021 IEEE International Conference on Robotics
  and Automation (ICRA)}.\hskip 1em plus 0.5em minus 0.4em\relax IEEE, 2021,
  pp. 1803--1809.

\bibitem{Zhang2011Datadrivenrob}
H.~Zhang, L.~Cui, X.~Zhang, and Y.~Luo, ``Data-driven robust approximate
  optimal tracking control for unknown general nonlinear systems using adaptive
  dynamic programming method,'' \emph{IEEE Transactions on Neural Networks},
  vol.~22, no.~12, pp. 2226--2236, 2011.

\bibitem{vamvoudakis2015asymptotically}
K.~G. Vamvoudakis, M.~F. Miranda, and J.~P. Hespanha, ``Asymptotically stable
  adaptive--optimal control algorithm with saturating actuators and relaxed
  persistence of excitation,'' \emph{IEEE transactions on neural networks and
  learning systems}, vol.~27, no.~11, pp. 2386--2398, 2015.

\bibitem{Kamalapurkar2018Reinforcementl}
R.~Kamalapurkar, P.~Walters, J.~Rosenfeld, and W.~E. Dixon, \emph{Reinforcement
  learning for optimal feedback control: A Lyapunov-based approach}.\hskip 1em
  plus 0.5em minus 0.4em\relax Springer, 2018.

\bibitem{Osinenko2022Reinforcementl}
\BIBentryALTinterwordspacing
P.~Osinenko, D.~Dobriborsci, and W.~Aumer, ``Reinforcement learning with
  guarantees: a review,'' \emph{IFAC-PapersOnLine}, vol.~55, no.~15, pp.
  123--128, 2022, presented at IFAC Conference on Intelligent Control and
  Automation Sciences. [Online]. Available:
  \url{https://www.sciencedirect.com/science/article/pii/S2405896322010308}
\BIBentrySTDinterwordspacing

\bibitem{Choi2020Reinforcementl}
J.~Choi, F.~Castaneda, C.~J. Tomlin, and K.~Sreenath, ``Reinforcement learning
  for safety-critical control under model uncertainty, using control lyapunov
  functions and control barrier functions,'' \emph{arXiv preprint
  arXiv:2004.07584}, 2020.

\bibitem{Cheng2019Endendsafe}
R.~Cheng, G.~Orosz, R.~M. Murray, and J.~W. Burdick, ``End-to-end safe
  reinforcement learning through barrier functions for safety-critical
  continuous control tasks,'' in \emph{Proceedings of the AAAI Conference on
  Artificial Intelligence}, vol.~33, no.~01, 2019, pp. 3387--3395.

\bibitem{Khasminskii2011StochasticStab}
R.~Khasminskii and G.~Milstein, \emph{Stochastic Stability of Differential
  Equations}, ser. Stochastic Modelling and Applied Probability.\hskip 1em plus
  0.5em minus 0.4em\relax Springer, 2011.

\bibitem{Bhasin2013novelactorcri}
S.~Bhasin, R.~Kamalapurkar, M.~Johnson, K.~G. Vamvoudakis, F.~L. Lewis, and
  W.~E. Dixon, ``A novel actor-critic-identifier architecture for approximate
  optimal control of uncertain nonlinear systems,'' \emph{Automatica}, vol.~49,
  no.~1, pp. 82--92, 2013.

\bibitem{Vamvoudakis2017Qlearningcont}
K.~G. Vamvoudakis, ``Q-learning for continuous-time linear systems: A
  model-free infinite horizon optimal control approach,'' \emph{Syst. Control
  Lett.}, vol. 100, pp. 14--20, 2017.

\bibitem{Vrabie2012OptimalAdaptiv}
D.~Vrabie, K.~G. Vamvoudakis, and F.~L. Lewis, \emph{Optimal Adaptive Control
  and Differential Games by Reinforcement Learning Principles}.\hskip 1em plus
  0.5em minus 0.4em\relax Institution of Engineering and Technology, 2012.

\bibitem{Plisnier2023TransferringMu}
H.~Plisnier, D.~Steckelmacher, J.~Willems, B.~Depraetere, and A.~Now{\'e},
  ``Transferring multiple policies to hotstart reinforcement learning in an air
  compressor management problem,'' \emph{arXiv preprint arXiv:2301.12820},
  2023.

\bibitem{Jiang2002converseLyapun}
Z.-P. Jiang and Y.~Wang, ``A converse lyapunov theorem for discrete-time
  systems with disturbances,'' \emph{Systems {\&} Control Letters}, vol.~45,
  no.~1, pp. 49--58, jan 2002.

\bibitem{Sontag1998Commentsintegr}
E.~D. Sontag, ``Comments on integral variants of {ISS},'' \emph{Systems and
  Control Letters}, vol.~34, no. 1-2, pp. 93--100, may 1998.

\bibitem{Osinenko2023actorcriticfr}
\BIBentryALTinterwordspacing
P.~Osinenko, G.~Yaremenko, G.~Malaniia, and A.~Bolychev, ``An actor-critic
  framework for online control with environment stability guarantee,''
  \emph{IEEE Access}, 2023. [Online]. Available:
  \url{https://ieeexplore.ieee.org/document/10223230}
\BIBentrySTDinterwordspacing

\bibitem{Astolfi1995Exponentialsta}
A.~Astolfi, ``Exponential stabilization of nonholonomic systems via
  discontinuous control,'' \emph{IFAC Proceedings Volumes}, vol.~28, no.~14,
  pp. 661--666, 1995.

\end{thebibliography}

\end{document}